\documentclass{article}
\usepackage{spconf,amsmath,amssymb,graphicx,hyperref}
\usepackage{stfloats}
\usepackage{arydshln}
\usepackage{graphicx}
\usepackage{booktabs}
\usepackage{makecell}
\usepackage{cite}  
\title{BEAP-AGENT: BACKTRACKABLE EXECUTION AND ADAPTIVE PLANNING FOR GUI AGENTS}
\name{
Ziyu Lu$^{1,2}$,
Tengjin Weng$^{1,2}$,
Yiying Yang$^{1}$,
Yuhang Zhao$^{1}$,
Xinxin Huang$^{3}$\textsuperscript{*},
Wenhao Jiang$^{1}$\textsuperscript{*}\thanks{*Corresponding authors.}
}

\address{$^{1}$ Guangdong Laboratory of Artificial Intelligence and Digital Economy (SZ), China \\
        $^{2}$ Shenzhen University, China \\
        $^{3}$ Guangdong University of Technology, China}

\begin{document}
\ninept
\maketitle
\begin{abstract}
GUI agents are designed to automate repetitive tasks and enhance productivity. However, existing GUI agents struggle to recover once they follow an incorrect exploration path, often leading to task failure. In this work, we model GUI task execution as a DFS process and propose BEAP-Agent, a DFS-based framework that supports long-range, multi-level state backtracking with dynamic task tracking and updating. The framework consists of three collaborative components: Planner, Executor, and Tracker. Together, they enable effective task exploration and execution. BEAP-Agent fills the gap in systematic backtracking mechanisms for GUI agents, offering a systematic solution for long-horizon task exploration. We conducted a systematic evaluation on the OSWorld benchmark, where BEAP-Agent achieved an accuracy of 28.2\%, validating the effectiveness of the proposed method.
\end{abstract}
\section{Introduction}
\label{sec:intro}

The primary objective of a Graphical User Interface (GUI) agent is to complete tasks by emulating human interactions with graphical interfaces \cite{10.24963/ijcai.2024/711, Hong2023CogAgentAV}. GUI agents are applicable to diverse scenarios, including automated testing, software workflows, information extraction, and the management of complex task pipelines \cite{zhang2025large}. Unlike traditional script-based automation, GUI agents obviate the need for task-specific adaptations, thereby providing greater robustness and flexibility in complex and dynamically changing environments \cite{wang2024oscar}.

In the early stages of GUI agent development, insufficient grounding ability was the primary cause of task failures. Due to the lack of deployable grounding mechanisms, models primarily relied on auxiliary structures such as HTML trees or accessibility trees \cite{wang2024oscar, wan2024omniparser,nguyen-etal-2025-gui, zhang-etal-2025-ufo, zhang2025ufo2}. However, these approaches exhibited limited generalizability across diverse environments. In the past year, many advanced vision-language models (VLMs) \cite{wang2024qwen2,wu2024atlas,bai2025qwen2,qin2025ui,xie2025scaling} have achieved significant improvements in grounding capabilities, with some demonstrating strong grounding performance relying solely on screen screenshots. As a result, an increasing number of GUI agents are shifting toward purely vision-based approaches built on screenshots, which also reduces the development difficulty of GUI agents.

At present, weak planning ability has become the main reason for task failures in GUI agents\cite{agashe2025agent, yang2025gta1}. For many GUI tasks, models lack precise and detailed knowledge to guarantee perfect execution. In practice, models often rely on approximate reasoning from similar tasks, producing high-level but imprecise plans that merely indicate a general direction \cite{yao2023react}. Consequently, an exploratory mechanism capable of supporting multiple rounds of trial-and-error is required to facilitate task completion.

\begin{figure}[t]
\centering
\includegraphics[width=0.9\columnwidth]{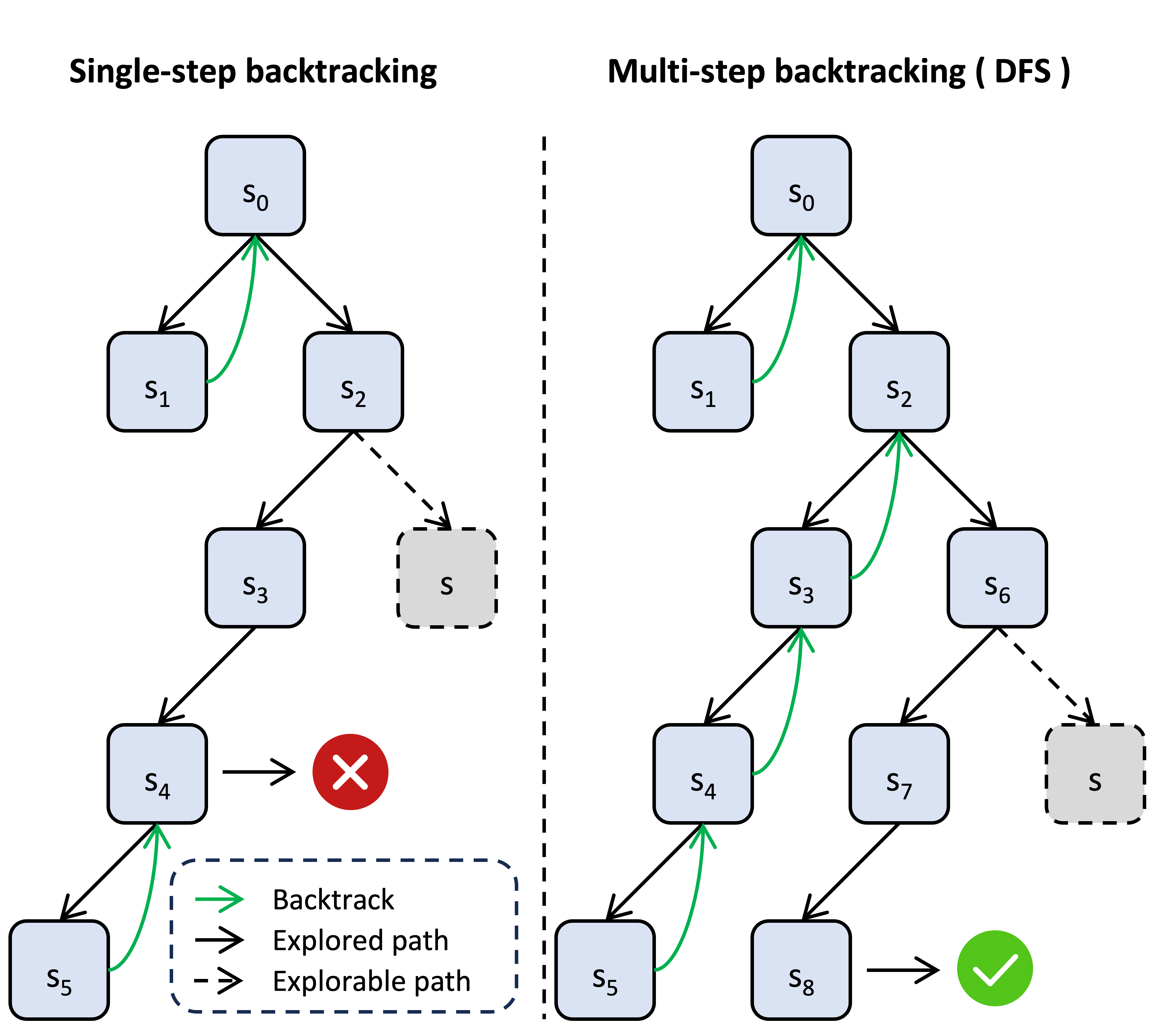}
\caption{Comparison of GUI exploration trajectories. The left side illustrates single-step backtracking, which fails to recover from error. The right side demonstrates multi-step, long-distance backtracking based on DFS modeling, which successfully completes the task. Dashed states denote unexplored yet explorable regions of the search space.}
\label{fig:tree}
\end{figure}

During the exploration process, current GUI agents may execute certain actions that lead to irreversible or suboptimal states. For example, in tasks involving multi-step form filling or web navigation, a single incorrect action may prevent the agent from returning to a valid state without backtracking, ultimately breaking the entire task chain and resulting in failure. BackTrackAgent \cite{wu2025backtrackagent}  was the first to highlight the importance of backtracking, treating the validity and task contribution of each action as evaluation criteria. After executing an action, BackTrackAgent verifies whether it passes all checks; if so, the agent continues exploration, otherwise it backtracks to the pre-action state and attempts an alternative. This represents a single-step backtracking mechanism, as illustrated in Fig.~\ref{fig:tree}. On many tasks, BackTrackAgent has demonstrated the effectiveness of this design.

However, our experimental observations reveal that when the agent recognizes an error, it is often not caused by the immediately preceding step. In most cases, the deviation from the correct trajectory only becomes apparent after progressing several steps, at which point it becomes clear that the path is entirely incorrect. This phenomenon arises because rewards in GUI tasks are highly sparse \cite{lu2025arpo, lian2025ui}, and the validity of many tasks can only be verified after executing multiple steps. Consequently, backtracking based solely on single-step judgments is ineffective in most cases, as the early actions typically provide no meaningful feedback.

In addition, relying exclusively on backtracking still presents challenges: the agent may prematurely terminate a task, omit critical operations, or fall into repetitive execution loops. Such problems stem not only from single-step action errors but also from misalignments between task progress and the current plan.

To address these issues, we formulate GUI task execution as a state-space tree search based on depth-first exploration (DFS). This formulation enables the agent to perform not only local exploration around the current step but also long-range backtracking to earlier historical states for deeper re-planning and exploration, thereby supporting long-horizon and multi-level recovery. Fig.~\ref{fig:tree} illustrates a comparison of the two exploration paradigms. Furthermore, we introduce BEAP-Agent, a DFS-based agent equipped with backtracking and dynamic task-tracking capabilities. The framework consists of three core components: Planner, Executor, and Tracker. The Planner generates task plans composed of multiple subtasks at initialization or after backtracking, guiding the exploration direction of BEAP-Agent. The Executor grounds these plans into executable actions for device interaction. The Tracker updates the subtask statuses within the plan while simultaneously determining the global execution status. The main contributions of this work are summarized as follows:
\begin{itemize}
    \item We formulate GUI task execution as a DFS–based state-space exploration problem, enabling long-horizon and multi-level backtracking that better aligns with the inherent sparsity of GUI rewards.
    \item We introduce BEAP-Agent, a novel framework that combines DFS exploration, backtracking, and dynamic task tracking. This design allows the agent to perform deep re-planning while maintaining consistency between task progress and planning.
    \item We conduct extensive experiments on GUI benchmarks, demonstrating that BEAP-Agent significantly outperforms existing baselines in task success rate.
\end{itemize}

\section{Methodology}
\label{sec:methodology}

\subsection{Backtracking Mechanism under DFS Modeling}

We model the GUI environment as a state space $S$, where each state $s \in S$ corresponds to a unique page configuration. Each action $a \in A(s)$ represents an operation executable in state $s$. The environment dynamics are defined by the transition function $T: S \times A \to S$, such that executing action $a$ in state $s$ produces a successor state $s' = T(s, a)$.

To emulate the \textit{trial-and-error, backtracking, and re-exploration} behaviors observed in human interactions with GUI environments, we formulate GUI task execution as a tree search problem in the state space. Each node of the search tree $\mathcal{T}$ corresponds to a state $s$, and each directed edge $(s, a, s’)$ represents executing action $a$ in $s$ and transitioning to successor state $s'$. Formally, the search tree is defined as:
\[
\mathcal{T} = \left\{ (s, a, s') \;\middle|\; s' = T(s, a),\; s \in S,\; a \in A(s) \right\}.
\]
For each state $s$, the set of unexplored paths is defined as:
\[
U(s) = \left\{ a \in A(s) \;\middle|\; (s, a, T(s,a)) \notin Z \right\},
\]
where $Z$ denotes the set of already explored state--action transitions. If $U(s) = \varnothing$, it indicates that all branches of state $s$ have been exhausted.

We adopt DFS as the exploration strategy. At a given state $s$, if there exists an unexplored path $a \in U(s)$, the agent executes $a$, leading to a new state $s' = T(s, a)$. If $U(s) = \varnothing$, backtracking is triggered, and the agent moves upward along the search path to the nearest ancestor node that still has unexplored paths. During this process, the failed exploration path is recorded to prevent redundant exploration, ensuring that $(s, a, s') \notin U(s)$ for all future steps.

The overall exploration process is illustrated in Fig.~\ref{fig:tree}. Formally, it can be expressed as:
\[
\text{DFS}(s) =
\begin{cases}
\text{FINISH}, & \text{if task is completed} \\[6pt]
\text{DFS}(T(s, a)), & \text{if } \exists a \in U(s) \\[6pt]
\text{Backtrack}(s), & \text{if } U(s) = \varnothing.
\end{cases}
\]
Thus, the execution of a GUI task can be viewed as a trajectory over the state--action search tree guided by DFS. The process terminates once the task is successfully completed or when the entire search space has been explored. This modeling provides a unified and rigorous framework for integrating planning, execution, backtracking, and recovery within GUI agents.

\subsection{BEAP-Agent}

\begin{figure*}[!t]
    \centering
    \includegraphics[width=0.9\textwidth]{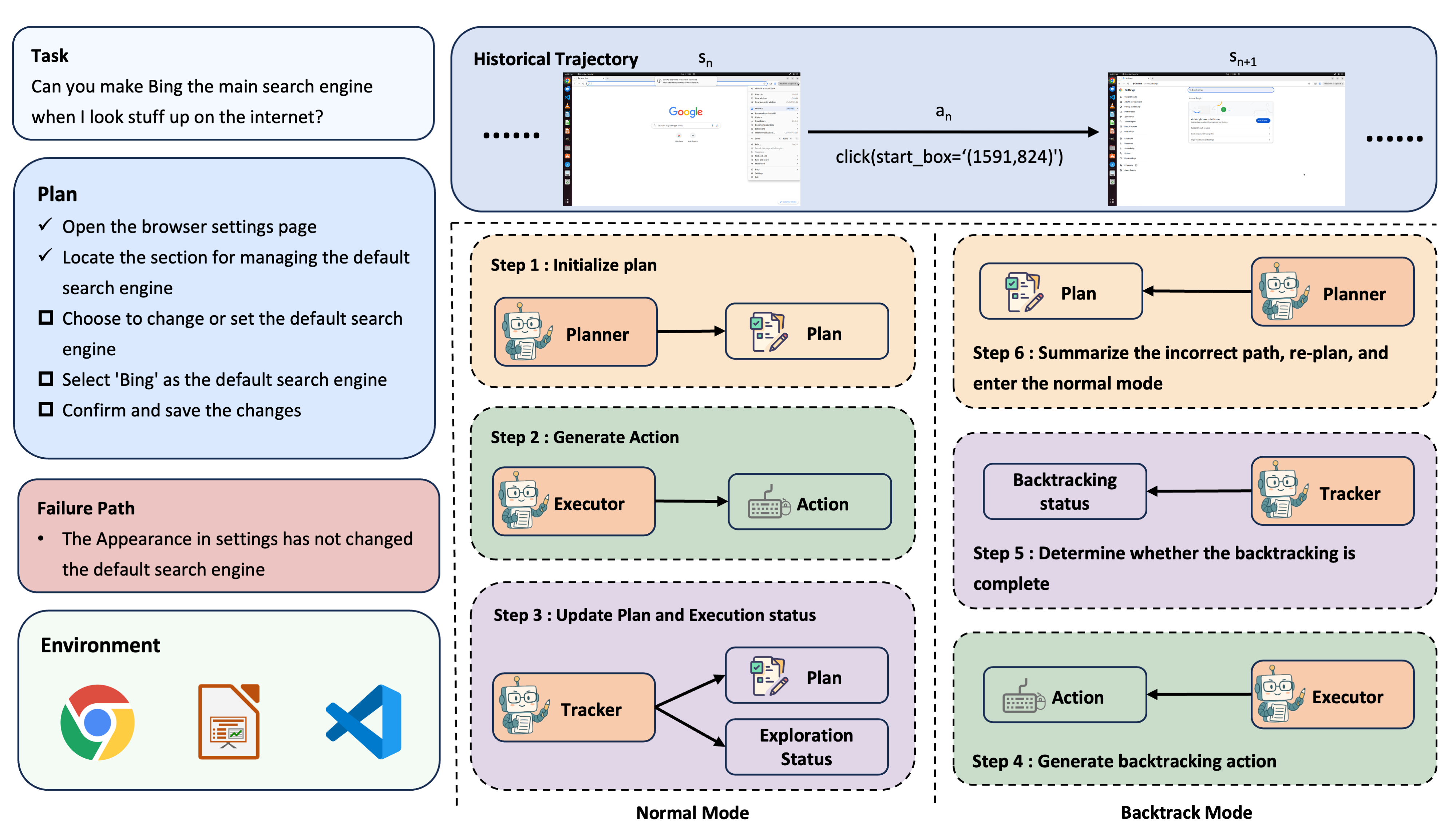}
    \caption{The overview of BEAP-Agent. The lower right part is the BEAP-Agent pipeline, and the rest is the data display of each module. Each module is different from normal mode and backtrack mode, where Tracker is responsible for switching between modes. }
    \label{fig:overview}
\end{figure*}

BEAP-Agent consists of three core modules: \textbf{Planner}, \textbf{Executor}, and \textbf{Tracker}. The overall execution workflow of BEAP-Agent is illustrated in Fig.~\ref{fig:overview}:

\begin{enumerate}
    \item At the beginning of a task, the Planner generates a task plan consisting of multiple subtasks.  
    \item Based on the current page state, task description, plan list, and historical trajectory, the Executor generates action commands to interact with the environment.  
    \item After executing the previous action, the Tracker updates the plan list according to the latest page state (e.g., updating the completion status of subtasks and adapting the remaining subtasks to better align with the actual environment). At the same time, the Tracker determines the exploration status: whether to terminate execution, continue exploration, or trigger backtracking. If continuation is chosen, the process returns to Step~2, if backtracking is triggered, the process moves to Step~4.  
    \item In backtracking mode, the Executor first generates actions that realize backtracking based on the historical trajectory. The Tracker then verifies whether backtracking has succeeded. If successful, the erroneous path is recorded, a new plan is generated, and the process returns to Step~2 in normal mode. If unsuccessful, the backtracking attempt is repeated.  
\end{enumerate}

\noindent\textbf{Planner.}  
Given a task $X$, the current page state $s$, and the set of explored failure paths $F$, 
the Planner outputs an initial plan or a revised plan that avoids failed paths:
\[
\text{Planner} : (s, X, F) \;\;\rightarrow\;\; P = \{(subtask_i, status_i)\}_{i=1}^n,
\]
where $status_i \in \{\texttt{PENDING}, \texttt{COMPLETED}\}$ indicates the execution status of the $subtask_i$. At initialization, all subtasks are assigned the status \texttt{PENDING}. 
In the case of plan regeneration, the statuses of completed subtasks are preserved, 
while the unfinished subtasks are revised by incorporating information from the previously explored failure paths.

\medskip
\noindent\textbf{Executor.}  
The Executor grounds plans into executable operations:
\[
\text{Executor} : (s, X, P, H) \;\;\rightarrow\;\; (a, s'),
\]
where $H$ denotes the historical trajectory, and $a$ is a well-defined primitive action executable within the environment. This module drives the state transition $s \rightarrow s'$. When in backtracking mode, Executor will use the historical trajectory to generate state backtracking actions.

\medskip
\noindent\textbf{Tracker.}  
After the Executor performs an action, the Tracker updates the plan and global execution status:  
\[
\text{Tracker}: (s, X, P, H, F) \;\;\rightarrow\;\;
\begin{cases}
(P', E), & \text{if normal mode} \\[2pt]
E_{\text{back}}, & \text{if backtrack mode},
\end{cases}
\]
where in normal mode, the Tracker outputs an updated plan $P'$ and the global execution status 
$E \in \{\texttt{CONTINUE}, \texttt{BACKTRACK}, \texttt{FAIL}, \allowbreak \texttt{DONE}\}$.  
The Tracker updates each subtask’s status from \texttt{PENDING} to \texttt{COMPLETED}, and adjusts incomplete subtasks to better fit the actual environment.  
At the same time, it determines the global execution status: \texttt{CONTINUE} means proceeding with the current plan, 
\texttt{BACKTRACK} means exploration is blocked and backtracking is required, 
\texttt{FAIL} means the task cannot be completed, and \texttt{DONE} means the task is completed.  

In backtrack mode, the Tracker is responsible for evaluating the outcome of the backtracking attempt. It produces an output $E_{\text{back}} \in \{\texttt{RECOVERED}, \texttt{NOT\_RECOVERED}\}$, indicating whether the process successfully returns the agent to a valid state. If recovery is achieved, the agent then switches back to normal execution mode.  

\section{Experiments}
\label{sec:experiments}

\subsection{Experimental Setup}
We evaluate our approach on the OSWorld \cite{xie2024osworld} benchmark, which comprises 369 real-world desktop tasks. These tasks span a wide range of categories, including operating system operations, office productivity, daily applications, professional software, and multi-application workflows. OSWorld provides a controlled execution environment for interactive agent learning, where task initialization and evaluation are conducted by executing actions on real-world operating systems through virtual machine techniques.  

To minimize external confounding factors and to specifically assess the effectiveness of backtracking in GUI tasks, we restrict the experimental setting to using only screenshots as input and four common human interaction actions as output: mouse click, mouse drag, mouse scroll, and keyboard input. No additional external tool calls are allowed, and all interactions are executed via the PyAutoGUI library. The maximum number of interaction steps is set to 50. We employ GPT-4o as the underlying model for both the Planner and Tracker, while UI-TARS-1.5-7B \cite{qin2025ui} is used as the Executor.

\subsection{Main Result}

\begin{table}[t]
\centering
\fontsize{9}{11}\selectfont
\begin{tabular}{@{}l l c c@{}}
\toprule
\textbf{Framework} & \textbf{Model} & \textbf{Steps} & \textbf{Acc} \\
\midrule
Agent S2 \cite{agashe2025agent} & GPT-4o + UI-TARS-72B-DPO & 50 & 26.6 \\
JEDI \cite{xie2025scaling} & GPT-4o + JEDI-7B & 50 & 25.0 \\
UI-TARS \cite{qin2025ui} & UI-TARS-1.5-7B & 50 & 24.0\textsuperscript{$\dag$} \\
AGUVIS \cite{xu2024aguvis} & GPT-4o + AGUVIS-72B & -- & 17.0 \\
Qwen2.5 \cite{bai2025qwen2} & Qwen2.5-vl-72B & -- & 8.8 \\
OpenAI & GPT-4o & -- & 5.0 \\
\addlinespace[2pt]
\cdashline{1-4}
\addlinespace[2pt]
\textbf{BEAP-Agent} & \textbf{GPT-4o + UI-TARS-1.5-7B} & \textbf{50} & \textbf{28.2} \\
\quad - w/o Backtrack & GPT-4o + UI-TARS-1.5-7B & 50 & 26.3 \\
\quad - w/o Tracker & GPT-4o + UI-TARS-1.5-7B & 50 & 23.6 \\
\midrule
\multicolumn{4}{l}{\textbf{Backtracking metrics}} \\
\multicolumn{2}{l}{\quad $\bullet$ Backtracking Task Rate} & \multicolumn{2}{c}{35.8\%} \\
\multicolumn{2}{l}{\quad $\bullet$ Backtrack Success Rate} & \multicolumn{2}{c}{65.5\%} \\
\multicolumn{2}{l}{\quad $\bullet$ Average Backtrack Steps} & \multicolumn{2}{c}{2.72} \\
\bottomrule
\end{tabular}
\caption{Performance comparison on OSWorld. For BEAP-Agent, we additionally report backtracking metrics. 
$\dagger$ indicates results from our reproduction.}
\label{tab:main-results}
\end{table}

Table~\ref{tab:main-results} presents the performance of BEAP-Agent on OSWorld \cite{xie2024osworld}. The results show that BEAP-Agent outperforms all existing methods. Under the same step limit, it improves accuracy by 17.5\% over the baseline method built on UI-TARS-1.5-7B \cite{qin2025ui}. This gain can be attributed to BEAP-Agent's unique backtracking mechanism and dynamic task update strategy, where the Planner incorporates external knowledge and the Tracker performs context-aware task updates and error backtracking, leading to superior performance. Agent S2 \cite{agashe2025agent} represents a strong GUI agent, yet BEAP-Agent still achieves a 6\% improvement in accuracy. JEDI \cite{xie2025scaling} adopts a basic Planner+Executor architecture, where at each step the Planner generates the next instruction and the Executor executes it. BEAP-Agent surpasses JEDI by 12.8\%. This shows that BEAP-Agent does not rely solely on GPT-4o for the introduction of external knowledge, and the agent exploration strategy plays a certain role.

\subsection{Ablation Study}

We also conduct an ablation study to validate the effectiveness of BEAP-Agent’s components. In the Table \ref{tab:main-results}, ``w/o Backtrack'' indicates the removal of the backtracking action, and ``w/o Tracker'' indicates the removal of the Tracker. This means that the agent generates a plan without maintaining any state and does not perform any plan updates afterwards.

As shown in the Table \ref{tab:main-results}, the complete BEAP-Agent achieves an accuracy of 28.2\% under 50-step testing. When the backtracking mechanism is removed, the accuracy drops to 26.3\%, indicating that backtracking allows the agent to effectively return to previous states after failed explorations and attempt alternative branches, thereby avoiding getting stuck in local dead-ends. This demonstrates that backtracking is one of the key factors for improving task success rates.

When the Tracker is removed, the accuracy declines even more significantly, reaching only 23.6\%, which is 4.6\% lower than the complete agent. Although the Planner still provides task completion knowledge, the absence of the Tracker prevents the agent from recovering from failed explorations, updating task progress, and adaptively modifying its plan. This leads the agent to deviate from the intended plan and ``spin its wheels'' in place, ultimately resulting in decreased performance.

\subsection{Experimental Analysis}

\begin{figure}[t]
\centering
\includegraphics[width=\columnwidth]{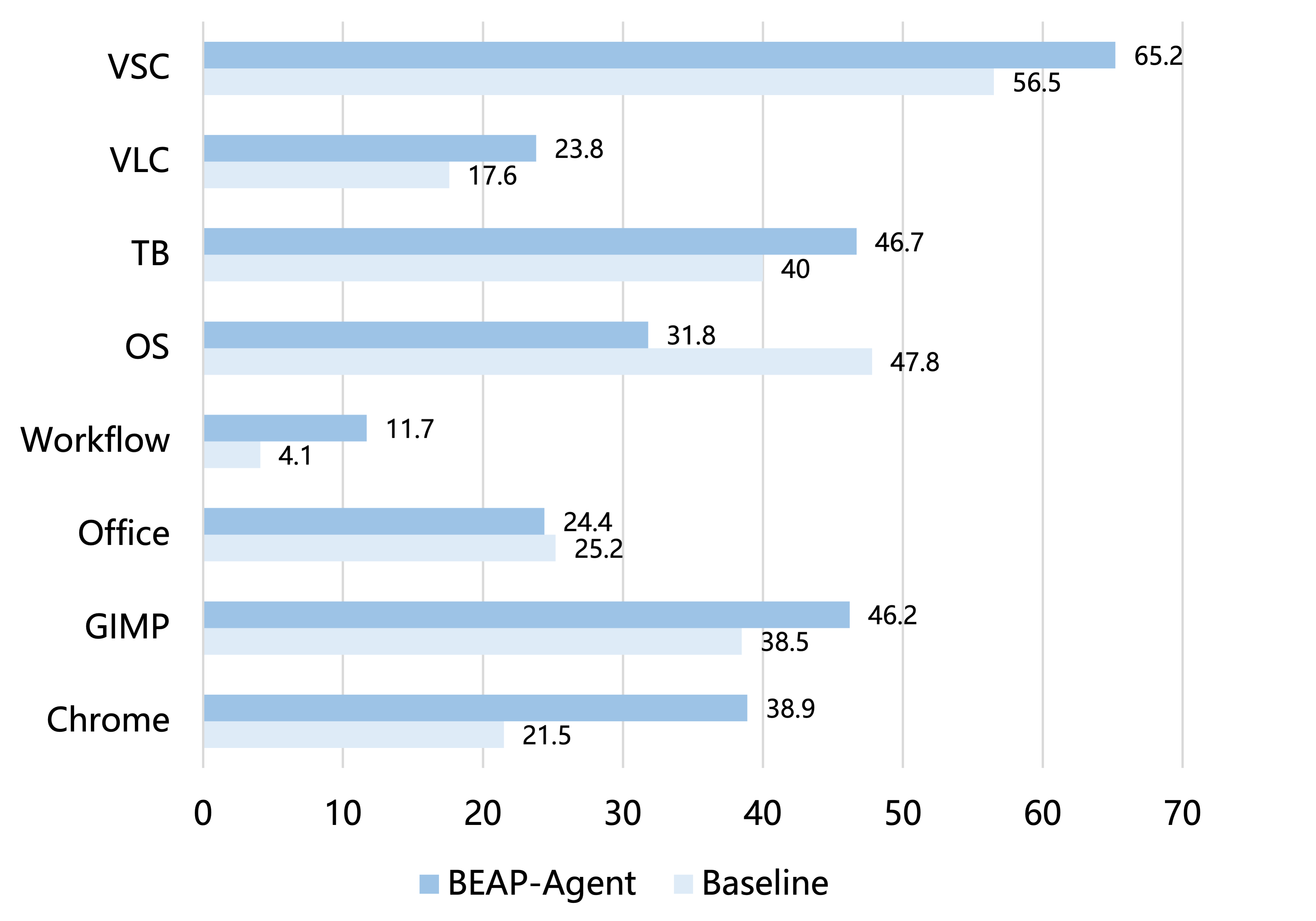}
\caption{Accuracy comparison across domains between BEAP-Agent and baseline.}
\label{fig:domain-result}
\end{figure}

We first conducted a systematic evaluation of the backtracking mechanism in BEAP-Agent, as shown in the Table \ref{tab:main-results}. Results indicate that 35.8\% of tasks triggered backtracking, and among these attempts, the agent successfully recovered to a valid state with a success rate of 65.5\%, demonstrating the robustness of the mechanism. Furthermore, the average number of steps required per backtracking attempt was 2.72, suggesting that the agent can efficiently recover from errors at relatively low cost.

As shown in Fig.~\ref{fig:domain-result}, BEAP-Agent outperforms the baseline across most domains. The advantage is particularly pronounced in domains such as chrome and workflow, where page states undergo more significant changes. In fact, the backtracking success rate in the chrome domain exceeds 80\%. This observation suggests that for tasks with pronounced state transitions, BEAP-Agent benefits more from the richer execution feedback available, enabling more effective task progress updates, exploration attempts, and backtracking. Backtracking actions share the same executor interface as normal actions, incurring no additional execution overhead. BEAP-Agent maintains state snapshots using a stack-based mechanism with sliding-window or checkpoint control, keeping memory overhead bounded and practical. Overall, these findings validate the effectiveness of BEAP-Agent in handling real-world desktop tasks.

\section{Conclusion}
\label{sec:conclusion}

In this study, we model GUI task execution as a DFS-based state-space exploration problem and propose BEAP-Agent. The framework employs three core components—Planner, Executor, and Tracker—to enable long-range, multi-level state backtracking and dynamic task tracking, allowing the agent to explore more robustly and efficiently in complex interactive environments. On the OSWorld benchmark, BEAP-Agent achieves a task success rate of 28.2\%, representing a 17.5\% relative improvement over baseline methods, which validates the effectiveness of the proposed framework. Ablation studies further demonstrate the crucial role of the backtracking and task-tracking mechanisms in improving overall performance. In addition, we observe that even advanced closed-source models such as GPT-4o remain limited in fine-grained screen-detail perception and icon understanding, and can sometimes be outperformed by smaller, task-specific models. This motivates future research on agent architectures that better integrate the complementary strengths of different models for cooperative task execution.

\bibliographystyle{IEEEbib}
\bibliography{refs}

@article{xie2024osworld,
  title={Osworld: Benchmarking multimodal agents for open-ended tasks in real computer environments},
  author={Xie, Tianbao and Zhang, Danyang and Chen, Jixuan and Li, Xiaochuan and Zhao, Siheng and Cao, Ruisheng and Hua, Toh J and Cheng, Zhoujun and Shin, Dongchan and Lei, Fangyu and others},
  journal={Advances in Neural Information Processing Systems},
  volume={37},
  pages={52040--52094},
  year={2024}
}

@article{qin2025ui,
  title={Ui-tars: Pioneering automated gui interaction with native agents},
  author={Qin, Yujia and Ye, Yining and Fang, Junjie and Wang, Haoming and Liang, Shihao and Tian, Shizuo and Zhang, Junda and Li, Jiahao and Li, Yunxin and Huang, Shijue and others},
  journal={arXiv preprint arXiv:2501.12326},
  year={2025}
}

@inproceedings{
agashe2025agent,
title={Agent S2: A Compositional Generalist-Specialist Framework for Computer Use Agents},
author={Saaket Agashe and Kyle Wong and Vincent Tu and Jiachen Yang and Ang Li and Xin Eric Wang},
booktitle={Second Conference on Language Modeling},
year={2025},
url={https://openreview.net/forum?id=zg5is4GJ3R}
}

@article{xie2025scaling,
  title={Scaling Computer-Use Grounding via User Interface Decomposition and Synthesis},
  author={Xie, Tianbao and Deng, Jiaqi and Li, Xiaochuan and Yang, Junlin and Wu, Haoyuan and Chen, Jixuan and Hu, Wenjing and Wang, Xinyuan and Xu, Yuhui and Wang, Zekun and others},
  journal={arXiv preprint arXiv:2505.13227},
  year={2025}
}

@article{xu2024aguvis,
  title={Aguvis: Unified pure vision agents for autonomous gui interaction},
  author={Xu, Yiheng and Wang, Zekun and Wang, Junli and Lu, Dunjie and Xie, Tianbao and Saha, Amrita and Sahoo, Doyen and Yu, Tao and Xiong, Caiming},
  journal={arXiv preprint arXiv:2412.04454},
  year={2024}
}

@article{bai2025qwen2,
  title={Qwen2. 5-vl technical report},
  author={Bai, Shuai and Chen, Keqin and Liu, Xuejing and Wang, Jialin and Ge, Wenbin and Song, Sibo and Dang, Kai and Wang, Peng and Wang, Shijie and Tang, Jun and others},
  journal={arXiv preprint arXiv:2502.13923},
  year={2025}
}

@article{wu2025backtrackagent,
  title={BacktrackAgent: Enhancing GUI Agent with Error Detection and Backtracking Mechanism},
  author={Wu, Qinzhuo and Gao, Pengzhi and Liu, Wei and Luan, Jian},
  journal={arXiv preprint arXiv:2505.20660},
  year={2025}
}

@article{
zhang2025large,
title={Large Language Model-Brained {GUI} Agents: A Survey},
author={Chaoyun Zhang and Shilin He and Jiaxu Qian and Bowen Li and Liqun Li and Si Qin and Yu Kang and Minghua Ma and Guyue Liu and Qingwei Lin and Saravan Rajmohan and Dongmei Zhang and Qi Zhang},
journal={Transactions on Machine Learning Research},
issn={2835-8856},
year={2025},
url={https://openreview.net/forum?id=xChvYjvXTp},
note={}
}

@inproceedings{wan2024omniparser,
  title={Omniparser: A unified framework for text spotting key information extraction and table recognition},
  author={Wan, Jianqiang and Song, Sibo and Yu, Wenwen and Liu, Yuliang and Cheng, Wenqing and Huang, Fei and Bai, Xiang and Yao, Cong and Yang, Zhibo},
  booktitle={Proceedings of the IEEE/CVF conference on computer vision and pattern recognition},
  pages={15641--15653},
  year={2024}
}

@inproceedings{zhang-etal-2025-ufo,
    title = "{UFO}: A {UI}-Focused Agent for Windows {OS} Interaction",
    author = "Zhang, Chaoyun  and
      Li, Liqun  and
      He, Shilin  and
      Zhang, Xu  and
      Qiao, Bo  and
      Qin, Si  and
      Ma, Minghua  and
      Kang, Yu  and
      Lin, Qingwei  and
      Rajmohan, Saravan  and
      Zhang, Dongmei  and
      Zhang, Qi",
    editor = "Chiruzzo, Luis  and
      Ritter, Alan  and
      Wang, Lu",
    booktitle = "Proceedings of the 2025 Conference of the Nations of the Americas Chapter of the Association for Computational Linguistics: Human Language Technologies (Volume 1: Long Papers)",
    month = apr,
    year = "2025",
    address = "Albuquerque, New Mexico",
    publisher = "Association for Computational Linguistics",
    url = "https://aclanthology.org/2025.naacl-long.26/",
    doi = "10.18653/v1/2025.naacl-long.26",
    pages = "597--622",
    ISBN = "979-8-89176-189-6"
}

@article{zhang2025ufo2,
  title={Ufo2: The desktop agentos},
  author={Zhang, Chaoyun and Huang, He and Ni, Chiming and Mu, Jian and Qin, Si and He, Shilin and Wang, Lu and Yang, Fangkai and Zhao, Pu and Du, Chao and others},
  journal={arXiv preprint arXiv:2504.14603},
  year={2025}
}

@inproceedings{yao2023react,
  title={React: Synergizing reasoning and acting in language models},
  author={Yao, Shunyu and Zhao, Jeffrey and Yu, Dian and Du, Nan and Shafran, Izhak and Narasimhan, Karthik and Cao, Yuan},
  booktitle={International Conference on Learning Representations (ICLR)},
  year={2023}
}

@inproceedings{10.24963/ijcai.2024/711,
author = {Niu, Runliang and Li, Jindong and Wang, Shiqi and Fu, Yali and Hu, Xiyu and Leng, Xueyuan and Kong, He and Chang, Yi and Wang, Qi},
title = {ScreenAgent: a vision language model-driven computer control agent},
year = {2024},
isbn = {978-1-956792-04-1},
url = {https://doi.org/10.24963/ijcai.2024/711},
doi = {10.24963/ijcai.2024/711},
booktitle = {Proceedings of the Thirty-Third International Joint Conference on Artificial Intelligence},
articleno = {711},
numpages = {9},
location = {Jeju, Korea},
series = {IJCAI '24}
}

@article{Hong2023CogAgentAV,
  title={CogAgent: A Visual Language Model for GUI Agents},
  author={Wenyi Hong and Weihan Wang and Qingsong Lv and Jiazheng Xu and Wenmeng Yu and Junhui Ji and Yan Wang and Zihan Wang and Yuxiao Dong and Ming Ding and Jie Tang},
  journal={2024 IEEE/CVF Conference on Computer Vision and Pattern Recognition (CVPR)},
  year={2023},
  pages={14281-14290},
  url={https://api.semanticscholar.org/CorpusID:266210390}
}

@article{yang2025gta1,
  title={Gta1: Gui test-time scaling agent},
  author={Yang, Yan and Li, Dongxu and Dai, Yutong and Yang, Yuhao and Luo, Ziyang and Zhao, Zirui and Hu, Zhiyuan and Huang, Junzhe and Saha, Amrita and Chen, Zeyuan and others},
  journal={arXiv preprint arXiv:2507.05791},
  year={2025}
}

@inproceedings{nguyen-etal-2025-gui,
    title = "{GUI} Agents: A Survey",
    author = "Nguyen, Dang  and
      Chen, Jian  and
      Wang, Yu  and
      Wu, Gang  and
      Park, Namyong  and
      Hu, Zhengmian  and
      Lyu, Hanjia  and
      Wu, Junda  and
      Aponte, Ryan  and
      Xia, Yu  and
      Li, Xintong  and
      Shi, Jing  and
      Chen, Hongjie  and
      Lai, Viet Dac  and
      Xie, Zhouhang  and
      Kim, Sungchul  and
      Zhang, Ruiyi  and
      Yu, Tong  and
      Tanjim, Mehrab  and
      Ahmed, Nesreen K.  and
      Mathur, Puneet  and
      Yoon, Seunghyun  and
      Yao, Lina  and
      Kveton, Branislav  and
      Kil, Jihyung  and
      Nguyen, Thien Huu  and
      Bui, Trung  and
      Zhou, Tianyi  and
      Rossi, Ryan A.  and
      Dernoncourt, Franck",
    editor = "Che, Wanxiang  and
      Nabende, Joyce  and
      Shutova, Ekaterina  and
      Pilehvar, Mohammad Taher",
    booktitle = "Findings of the Association for Computational Linguistics: ACL 2025",
    month = jul,
    year = "2025",
    address = "Vienna, Austria",
    publisher = "Association for Computational Linguistics",
    url = "https://aclanthology.org/2025.findings-acl.1158/",
    doi = "10.18653/v1/2025.findings-acl.1158",
    pages = "22522--22538",
    ISBN = "979-8-89176-256-5"
}

@article{lu2025arpo,
  title={ARPO: End-to-End Policy Optimization for GUI Agents with Experience Replay},
  author={Lu, Fanbin and Zhong, Zhisheng and Liu, Shu and Fu, Chi-Wing and Jia, Jiaya},
  journal={arXiv preprint arXiv:2505.16282},
  year={2025}
}

@article{lian2025ui,
  title={UI-AGILE: Advancing GUI Agents with Effective Reinforcement Learning and Precise Inference-Time Grounding},
  author={Lian, Shuquan and Wu, Yuhang and Ma, Jia and Song, Zihan and Chen, Bingqi and Zheng, Xiawu and Li, Hui},
  journal={arXiv preprint arXiv:2507.22025},
  year={2025}
}

@article{wang2024qwen2,
  title={Qwen2-vl: Enhancing vision-language model's perception of the world at any resolution},
  author={Wang, Peng and Bai, Shuai and Tan, Sinan and Wang, Shijie and Fan, Zhihao and Bai, Jinze and Chen, Keqin and Liu, Xuejing and Wang, Jialin and Ge, Wenbin and others},
  journal={arXiv preprint arXiv:2409.12191},
  year={2024}
}

@article{wu2024atlas,
    title={OS-ATLAS: A Foundation Action Model for Generalist GUI Agents},
    author={Wu, Zhiyong and Wu, Zhenyu and Xu, Fangzhi and Wang, Yian and Sun, Qiushi and Jia, Chengyou and Cheng, Kanzhi and Ding, Zichen and Chen, Liheng and Liang, Paul Pu and others},
    journal={arXiv preprint arXiv:2410.23218},
    year={2024}
}

@article{wang2024oscar,
  title={Oscar: Operating system control via state-aware reasoning and re-planning},
  author={Wang, Xiaoqiang and Liu, Bang},
  journal={arXiv preprint arXiv:2410.18963},
  year={2024}
}

\end{document}